\documentclass[runningheads]{llncs}
\usepackage[T1]{fontenc}
\usepackage{graphicx}
\usepackage{amsmath}
\usepackage[table]{xcolor}
\usepackage{todonotes}
\newcommand{\proj}{\pi}

\begin{document}
%

\title{Querying Structured Data Through Natural Language Using Language Models}
%
%
\author{Valentin-Micu Hontan\inst{1} \and
Andrei-Alexandru Bunea\inst{1} \and
Nikolaos Dimitrios Tantaroudas\inst{2} \and
Dan-Matei Popovici\inst{1}} 
\authorrunning{Hontan et al.}
%
\institute{National University of Science and Technology POLITEHNICA Bucharest\\
\email{valentin.micu@stud.acs.upb.ro, andrei.bunea@stud.acs.pub.ro, matei.popovici@upb.ro} \and 
Institute of Communication and Computer Systems (ICCS), Athens, Greece\\
\email{nikolaos.tantaroudas@iccs.gr}
}
\maketitle              
\begin{abstract}
This paper presents an open-source methodology for allowing users to query structured, non-textual datasets through natural language. Unlike Retrieval-Augmented Generation (RAG), which struggles with numerical and highly structured information, our approach trains an LLM to generate executable queries. To support this capability, we introduce a principled pipeline for synthetic training-data generation, producing diverse question–answer pairs that capture both user intent and the semantics of the underlying dataset. We fine-tune a compact model—DeepSeek R1-Distill-8B—using QLoRA with 4-bit quantization, making the system suitable for deployment on commodity hardware.

We evaluate our approach on a dataset describing accessibility to essential services across Durangaldea, Spain. The fine-tuned model achieves high accuracy across monolingual, multilingual, and unseen-location scenarios, demonstrating both robust generalization and reliable query generation.  Our results highlight that small, domain-specific models can achieve high-precision for this task without relying on large proprietary LLMs, making this methodology suitable for resource-constrained environments and adaptable to broader multi-dataset systems.

\keywords{Natural Language querying \and tool-using LLMs \and model fine-tuning \and Large Language Models}
\end{abstract}
\section{Introduction}

The reasoning capabilities of language models are inherently constrained by the patterns and information from the data they have seen during training. To overcome this constraint, a substantial body of research focuses on augmenting language models with external knowledge sources. A widely adopted such method is Retrieval-Augmented Generation (RAG) \cite{lewis2020retrieval}.

In a typical industrial RAG pipeline, when the model is prompted with a query, an external retrieval system searches a text corpus and returns the top-$k$ most relevant documents. These retrieved fragments are then injected into the model’s context — often combined with a system prompt that provides instructions on how the information should be used - enabling the language model to produce an accurate, data-backed answer.

Similarity search in a RAG system operates by representing text fragments as vector embeddings in an $n$-dimensional space. A user query is encoded into the same space, and the system identifies the embeddings in the database that are most similar to it. Similarity is typically computed using the $\ell_2$ (Euclidean) distance, after which the plain-text associated with the nearest embeddings is retrieved. 

RAG has been successfully deployed in a wide range of applications, and most production-grade LLM-powered systems rely on variants of this approach. However, RAG is limited when a system must access specialized information that is not naturally expressed as unstructured text—for example, numerical data or time-series. Consider the following query: 

\begin{quote}\emph{Enumerate small towns that have good access to hospitals (in minutes by car) or to supermarkets (in minutes by bike).}
\end{quote}
Such questions require precise, structured, and often updated information that cannot be reliably retrieved through text alone. Given a dataset capable of providing the required information, we would like a system that can understand its structure, extract and reason about the relevant data, and use it in order to formulate a response.

This capability is already feasible for LLMs with hundreds of billions of parameters and context windows on the order of 100k tokens, as supported by many recent state-of-the-art commercial models. In such settings, the extensive context window allows developers to supply detailed descriptions of the dataset schema, access methods, API interfaces, and related metadata directly within the prompt. Data access formats have also been standardized through frameworks such as Anthropic’s Model Context Protocol (MCP) \cite{anthropic_mcp_2024}, an open protocol that enables AI systems to securely interact with external tools, data sources, and services in a structured and interoperable manner.

However, the practicality and deployment of such systems is constrained by their reliance on extremely large language models. This limitation is twofold. First, many approaches depend on closed-source LLMs—such as those provided by OpenAI, Anthropic or xAI’s Grok, or Google’s Gemini—which raises concerns regarding operational cost, data privacy, and control over the underlying infrastructure. Second, running open-source models at the hundred-billion– to trillion-parameter scale is generally infeasible for most organizations due to prohibitive hardware requirements.

Alternative strategies exist, such as generating SQL queries directly from natural language \cite{sql_generate,gao2023texttosqlempoweredlargelanguage}, but these methods are syntax or query-specific and tend to achieve strong performance only when paired with very large prediction models.

In this paper, we propose a fully open-source methodology for building LLM-based systems that can answer natural-language questions using specialized, non-textual data. Our approach introduces the following key contributions: (i) it relies on comparatively \textbf{small models}—including those in the 7-8B parameter range—while achieving high accuracy and (ii) supporting a workflow that can be \textbf{almost entirely automated}. Our approach is based on model fine-tuning, which may require specialized hardware, but remains (iii) \textbf{cost-effective} given the relatively small model sizes we target. We employ larger models, such as GPT-4, only during the training phase, not at deployment. Once trained, our system operates independently of any large or proprietary model.

We apply our methodology to a dataset describing accessibility conditions in the Durangaldea region of northern Spain. This dataset was collected and curated by the German Aerospace Center (DLR) as part of the FUTURAL (Future Rural) project, which aims to bring data-driven intelligence to rural communities. Our system is currently being piloted to support stakeholders in exploring the dataset and extracting actionable insights, thereby contributing to Quality of Life assessments in the Durangaldea region. We further argue that our approach can be integrated into a multi-dataset system and, when paired with recent highly efficient million-parameter architectures such as Tiny Recursive Models \cite{jolicoeurmartineau2025morerecursivereasoningtiny}, can be deployed on resource-constrained devices, including laptops.

The remainder of this paper is organized as follows: Section \ref{sec:problem} introduces the problem setting and outlines the overall methodology. Section \ref{sec:dataset} describes the construction of the dataset we used for training and evaluation. Section \ref{sec:finetuning} illustrates the model fine-tuning procedure. Section \ref{sec:evaluation} reports the evaluation metrics and experimental results.
Section \ref{sec:app-performance} analyzes the performance of the deployed application. Section \ref{sec:limitations} discusses limitations of the current approach and directions for future research.
Section \ref{sec:related-work} reviews related work, and Section \ref{sec:conclusion} concludes the paper.

\section{Problem Setting and Methodology}\label{sec:problem}

The DLR dataset we employ contains more than 100,000 records detailing geographic coordinates and travel times (walking, cycling, and driving) to hospitals, supermarkets, and pharmacies across Durangaldea, Biscay (Basque Country, northern Spain), covering over 240 km². It was developed for an ongoing study on how accessibility to essential services has changed in rural Durangaldea, examining the relationship between demographic dynamics, settlement patterns, and walkability.

Our goal is to build a natural-language interface to the dataset, enabling accessibility experts and other stakeholders to query it interactively. We aimed for a scalable, easily deployable solution that generalizes beyond the Durangaldea data, decouples the model from the dataset (which may evolve), and remains agnostic to query structure and syntax, assuming only that data access is available through a known query language or API. Additionally, we only considered a fully open, non-proprietary approach that does not rely on closed-source language models.

\begin{figure}[t] 
    \centering
    \includegraphics[trim=0 100 150 0, width=0.8\linewidth]{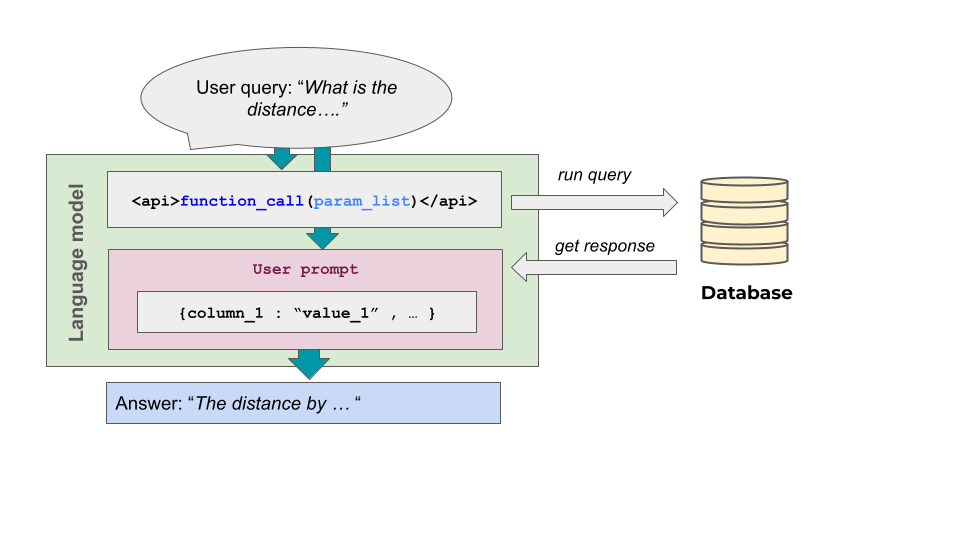}
    \caption{Overview of the model inference and query generation process}
    \label{fig:methodology}
\end{figure}

Retrieval-Augmented Generation proved inadequate for numerical information, as embeddings do not reliably encode structured quantitative data. We therefore explored a system in which the model generates executable queries, retrieves the corresponding data, and then reasons over the results. The overall functionality of our system is outlined in Figure~\ref{fig:methodology}. Instead of producing natural-language output directly, the model first generates a query based on the user question; the query is then executed; the returned data is fed back into the (user) prompt for answer generation. This approach, inspired by tool-use prompting methods such as those proposed in~\cite{yang2023gpt4toolsteachinglargelanguage}, requires an additional, detailed, system prompt describing the dataset schema and query syntax. While effective with very large models, our experiments showed that smaller models often produce syntactically incorrect or semantically irrelevant queries, limiting accuracy.

Early query-generation approaches such as~\cite{Schick2023ToolformerLM} rely on large training datasets to teach models when and how to generate queries. In such work, an external factual dataset is instrumented by removing selected pieces of knowledge and replacing them with query–answer examples (e.g., “What is the current population of Berlin?”). During training, the model learns to recognize when information is missing and a query is required. After such a query is generated, text generation is paused, the query is executed, and the returned result is inserted into the context before generation resumes.



This approach depends on the availability of high-quality training data containing diverse and accurate question–answer pairs. However, such datasets rarely exist for specialized or domain-specific databases, including the one considered in our work. To address this limitation, we propose a method for \textbf{synthetic dataset creation}, focusing on broad coverage of dataset values to support generalization, while ensuring rich linguistic variation.


Building an expert-level data retriever with LLMs is supported by various commercial solutions; however, our solution diverges from these approaches and is guided by two key constraints: (i) it must \textbf{run on consumer-grade hardware}. In this work we focus on the NVIDIA GeForce RTX 3090, a 24-GB Ampere GPU suitable for inference and light training which is significantly more affordable than datacenter-class GPUs such as the A100; and (ii) it must rely solely on \textbf{open-source models}. We use DeepSeek R1-Distill-8B~\cite{Guo2025DeepSeekR1}, a compact distilled variant of the DeepSeek-R1 family. Distillation transfers reasoning capabilities from a large model into a smaller one; in this case, the architecture is based on the 8-B Llama-3.1 Instruct model~\cite{noauthor_llama_2024}, fine-tuned on reasoning-oriented data. The resulting model is lightweight enough for deployment on the RTX 3090 while retaining strong problem-solving performance.

We fine-tune our model through the LoRA (Low-Rank Adaptation)~\cite{qlora} adapter framework. LoRA introduces trainable low-rank matrices into selected weight layers of a frozen base model, allowing the system to learn domain-specific tasks without modifying the original model parameters. Adapters operate as lightweight, modular extensions that can be loaded or unloaded dynamically, enabling efficient specialization while preserving the underlying model’s general linguistic and reasoning capabilities. This design makes training resource-efficient—only a small number of additional parameters are trained. At the same time deployment remains lightweight, since adapters require minimal memory and can be swapped in almost instant time during inference.

A central advantage of this architecture is its \textbf{extensibility}: a single base model can be paired with multiple adapters, each tailored to a different dataset or domain, thereby supporting multi-dataset systems without the need to duplicate the full model. We integrate the fine-tuned model into an application pipeline that combines geolocation services and a user-facing interface, enabling natural-language queries to be mapped onto structured queries and executed over the underlying dataset.

\section{Dataset creation}\label{sec:dataset}

Our objective was to construct a training dataset containing question–answer pairs that accurately capture both user intent and the information present in the underlying data. An additional challenge was that the dataset did not reference street addresses; instead, all locations were specified through geographic coordinates in the WGS-84~\cite{WGS84_TR8350_2} coordinate system.


Our dataset-generation procedure proceeds as follows. For each table $T$ available in the database, we consider all projections $\proj_{A_1,\ldots, A_i}(T)$ that capture semantically meaningful information. For every such projection, we construct a set of template questions designed to capture the types of information expressible from that attribute subset. Table~\ref{tab:projections} illustrates an example: a question instantiated from a template derived from a projection of the \emph{Hospitals} table containing the attributes \emph{Location} and \emph{Distance}.

\begin{table}[h]
\centering
\begin{tabular}{l r}
Projection & $\proj_{\textnormal{\tiny Location},\textnormal{\tiny Drive\_Dist}}(Hospitals)$ \\[6pt]
\hline
Template & What is the nearest \texttt{location\_type} from \texttt{location}? \\[6pt]
\hline
Question & What is the nearest hospital from Durango? \\[6pt]
\hline
\end{tabular}
\caption{Generating question from projections} \label{tab:projections}
\end{table}
For each identified projection, we use state of the art models (Gemini 2.5 Pro and DeepSeek R1) to generate templates that convey distinct semantic interpretations. We additionally include relevant superprojections—projections $p_j$ whose attribute sets strictly contain those of $p_i$ whenever they give rise to questions that reflect meaningfully different informational aspects. 
For instance, the question \emph{Is the nearest hospital closer by bike or on foot?} is generated from a template $p$ which is a superprojection of $\proj_{\textnormal{\tiny Location}}(Hospitals)$.

To ensure soundness, we manually inspect generated templates and filter-out invalid or duplicate templates. When instantiating each question template, we used the Overpass API~\cite{overpass_api} to extract a representative geographical sample consisting of 358 unique locations from the Durangaldea region.

For every instantiated question, we also generated a corresponding correct answer, which includes the associated database query used to compute it, as illustrated in Figure~\ref{fig:question-answer}.

\begin{figure}[h]
\centering
\begin{tabular}{p{0.25\textwidth} p{0.72\textwidth}}
\emph{Question} & What is the nearest hospital from Durango by drive? \\[6pt]
\emph{Correct answer} \newline \emph{with query} & {\scriptsize The closest hospital you can find is \texttt{<API>get\_closest\_distance\_time(category="hospital", mode="drive", location="Abadiño, Durango", metric\_to\_extract="distance") -> {"distance": 0.402, "time": 0.537}</API>} 0.402km away.}
\end{tabular}
\caption{Question-answer pair for model fine-tuning}\label{fig:question-answer}
\end{figure}


During inference, the model generates a call whenever it deems necessary. Once the call is produced, token generation is paused, the database query is invoked, and the returned value is inserted into the context. When generation resumes, the model has access to the complete call–response pair and can use it to formulate its final answer.
For this reason, the training data must contain both the question and the resulting answer, ensuring the model learns how to incorporate the query output into a coherent and correct final response.

Finally, to ensure sufficient linguistic variability, we employed state-of-the-art models (Gemini 2.5 Pro and DeepSeek R1) to generate syntactic paraphrases of the questions while preserving all location-specific information from the dataset. This step enables evaluation of the model’s robustness to paraphrasing and its ability to retain the parameters required for accurate querying.

The complete dataset thus created contains 44849 question answer pairs.


\section{Model fine-tuning}\label{sec:finetuning}

We fine-tuned the DeepSeek R1-Distill-8B model, which exhibits reasoning abilities exceeding those of typical models of comparable size. Effective query generation requires structured reasoning: recognizing when a query is necessary, identifying the required arguments, and mapping question semantics to the appropriate query. The model’s strong latent reasoning capacity makes it well suited for learning these decision patterns~\cite{deepseekr1}. 
Moreover, the model displays low variance in output structure relative to similarly sized alternatives, improving query generation reliability.

The model was fine-tuned on a single NVIDIA RTX 3090 GPU using the Transformer Reinforcement Learning library, specifically the SFTTrainer component, which provides an optimized framework for supervised fine-tuning of large language models. To reduce memory consumption and enable efficient training on commodity hardware, we employed 4-bit quantization~\cite{frantar2022gptq}, which compresses model weights while preserving core representational capacity. Fine-tuning was performed using QLoRA~\cite{qlora} adapters, allowing us to update only a small set of low-rank adaptation parameters: approximately 10M parameters were trained, rather than the full model.

Training was done with a batch size of 32 examples and a cosine learning-rate scheduler, chosen for its smooth decay properties that support stable convergence during fine-tuning. The model was trained for four epochs, each requiring several hours of compute time. To mitigate overfitting, we periodically evaluated performance on a held-out validation set and saved checkpoints corresponding to the lowest validation loss. In addition, we implemented an Early Stopping criterion that halted training when no further improvement was observed, ensuring computational efficiency and preventing unnecessary parameter updates.

\begin{figure}[t]
\centering
\includegraphics[width=0.85\textwidth]{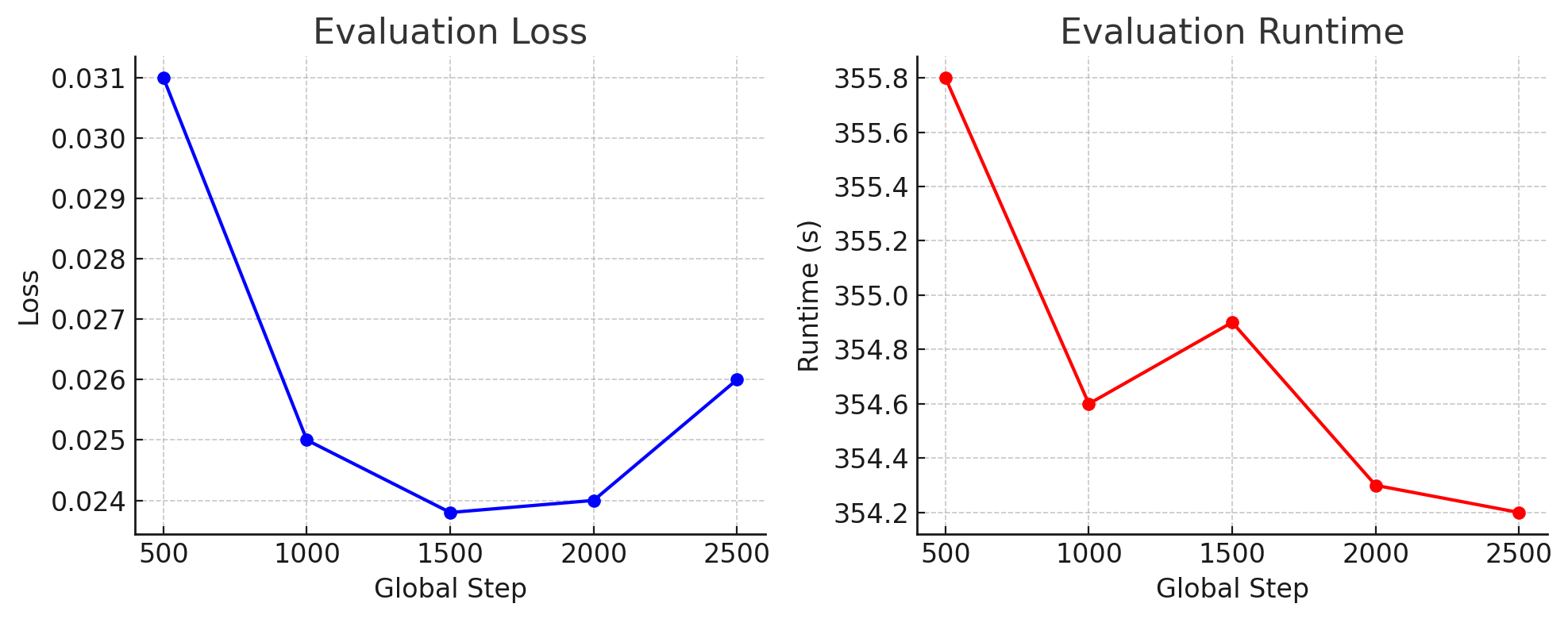}
\caption{Evaluation metrics during fine-tuning.}
\label{fig:eval_metrics}
\end{figure}

Figure~\ref{fig:eval_metrics} illustrates the evolution of the evaluation loss (left) and the evaluation runtime (right) across training steps. The loss curve shows a clear downward trend during the early stages of training, dropping from approximately 0.031 to 0.024 by step 1500. This indicates that the model rapidly learns the task structure during the initial epochs. After this point, the loss stabilizes, with only minor fluctuations; the slight increase observed at step 2500 suggests the onset of overfitting, consistent with the Early Stopping criterion applied during training.

The evaluation runtime remains effectively constant throughout training, with variations of less than two seconds across all checkpoints. This stability reflects the fixed computational footprint of QLoRA-based fine-tuning: because only a small set of adapter parameters is updated, the computational cost of forward passes remains unchanged. The absence of runtime drift confirms that neither model degradation nor GPU resource saturation occurred during training, further validating the reliability of the training setup.

The final model is publicly available~\footnote{\url{https://huggingface.co/valy3124/durangaldea-assistantFinalPD}}, allowing anyone to reproduce the results.

\section{Evaluation}\label{sec:evaluation}

\subsection{Metrics}

We evaluated system performance using two standard NLP metrics: ROUGE-L and BLEU-4. These metrics provide complementary assessments of structural alignment and syntactic precision. ROUGE-L utilizes the Longest Common Subsequence (LCS) to quantify the overlap between reference ($R$) and generated ($G$) tokens. By focusing on sequence order rather than strict contiguity, ROUGE-L effectively captures the preservation of key informational units—such as query arguments and factual components—even in the presence of paraphrasing. The metric is illustrated in Equation~a
where $LCS_{len}$ is the length of the longest common subsequence, and $|R|$ and $|G|$ represent the respective lengths of the reference and generated sequences. BLEU-4 measures similarity via contiguous $n$-gram overlaps (1- to 4-grams), making it highly sensitive to syntactic accuracy. This sensitivity is critical for query generation, where minor deviations in punctuation or token order can significantly degrade output validity. As shown in Equation~b, 
the score is calculated using the geometric mean of modified $n$-gram precisions $p_n$. We assign equal weights ($w_n = 1/4$) to each $n$-gram and apply a Brevity Penalty ($BP$) to penalize insufficiently long outputs. While BLEU-4 is less tolerant of paraphrasing than ROUGE-L, it serves as a rigorous proxy for the model's adherence to required syntactic structures.

\begin{equation*}
\begin{array}{cc}
\text{ROUGE-L} = \frac{2\,\mathrm{LCS}_{\mathrm{len}}}{|R| + |G|} \quad (a) & \text{BLEU-4} = BP \times \exp\left(\sum_{n=1}^{4} w_n \log p_n\right) \quad (b)\\
\end{array}
\end{equation*}


In practice, both metrics are computed for each generated response and then averaged across the evaluation set, providing different perspectives on semantic fidelity (ROUGE-L) and syntactic accuracy (BLEU-4).

Apart from BLEU-4 and ROUGE we also look at the percentage of exact matches ($EM$), namely the proportion of answers in which the generated query matches the reference exactly (Equation~\ref{eq:em}), with no missing parameters, incorrect values, misspellings or wrong parameter order.

\begin{equation}\label{eq:em}
\begin{array}{c}
\text{EMA} = \frac{N_{\text{exact}}}{N_{\text{total}}} \times 100\%
\end{array}
\end{equation}

\subsection{Evaluation results}

For evaluation, we reserved 2,800 queries from the training dataset as a held-out monolingual test set. In addition, we generated 500 multilingual queries using the same procedure described in Section~\ref{sec:dataset}. This evaluation design serves two purposes. First, it allows us to assess the model’s ability to generalize across languages and dialects. Although DeepSeek is pretrained on a multilingual corpus, our fine-tuning was performed exclusively in English; therefore, testing multilingual generalization provides insight into how much of the original linguistic capability is retained after adaptation. Second, multilingual support is of practical importance, as real users in the Durangaldea region may interact with the application in different spoken languages.

We also evaluated the model on entirely unseen geographic locations, isolating this scenario to measure its ability to extrapolate beyond the spatial distribution encountered during training. This is essential for determining the robustness of location-dependent reasoning and query formation.

Finally, we analyzed semantic variations of questions whose locations were seen during training and found that the model exhibits best performance on these. The complete set of evaluation results is shown in Table~\ref{tab:evaluation-results}.

\begin{table}[t]
\centering
\renewcommand{\arraystretch}{1.05}
\small
\begin{tabular}{|l|c|c|c|c|}
\hline
\textbf{Case / Subset} & \textbf{Size} & \textbf{Exact Match (\%)} & \textbf{BLEU-4} & \textbf{ROUGE-L} \\
\hline
\textbf{Unseen Locations} & 500 & 89.0 & 0.99 & 0.98 \\
\hline
\textbf{Semantic Variants} & 500 & \textbf{94.2} & 0.99 & 0.99 \\
\hline
\multicolumn{5}{|l|}{\textbf{Multilingual queries}} \\
\hspace{1em}Spanish              & 100 & 81.0 & 0.97 & 0.95 \\
\hspace{1em}Catalan              & 100 & 86.0 & 0.98 & 0.96 \\
\hspace{1em}Basque               & 100 & 24.0 & 0.81 & 0.71 \\
\hspace{1em}Galician             & 100 & 93.0 & 0.99 & 0.98 \\
\hspace{1em}French               & 100 & 64.0 & 0.95 & 0.91 \\
\hline
\textbf{Full dataset evaluation} & 3300 & 80.0 & 0.96 & 0.93 \\
\hline
\shortstack{\textbf{Full dataset evaluation} \\ (excluding multilingual cases)} & 2800 & 85.0 & 0.97 & 0.95 \\
\hline
\end{tabular}
\caption{Model accuracy}
\label{tab:evaluation-results}
\end{table}

Our model shows strong generalization to unseen locations, achieving an Exact Match of 89\% and near-perfect BLEU-4 and ROUGE-L scores. This indicates that the model reliably handles novel geographic inputs and correctly forms API calls even when encountering regions not present during fine-tuning.

For multilingual queries, performance varies substantially by language. Accuracy remains high for Spanish, Catalan, Galician, and French, reflecting the multilingual prior knowledge inherited from the DeepSeek pretraining. In contrast, performance on Basque is markedly lower, which is consistent with Basque’s relatively limited representation in most large-scale multilingual corpora. This suggests that while the fine-tuned model preserves substantial multilingual capability, its robustness depends strongly on the linguistic coverage of the underlying foundation model.

When evaluating the full dataset, including both monolingual and multilingual scenarios, the model achieves an Exact Match of 80\%. Restricting evaluation to monolingual English queries increases performance to 85\%, confirming that multilingual variability is the main source of accuracy degradation. BLEU-4 and ROUGE-L remain consistently high across settings, indicating that even when the model does not achieve exact matches, its predictions typically preserve the correct query structure and essential content.

We evaluated the performance of our proposed model against several state-of-the-art (SOTA) large language models (LLMs). To ensure a fair comparison, the SOTA models were evaluated in a \textbf{zero-shot setting} (no additional training or fine-tuning), using structured prompts that detailed the API schema, dataset characteristics, and query generation logic.
\begin{table}[t]
\centering
\begin{tabular}{lccccc}
\hline
\textbf{Model} & \textbf{Accuracy (\%)} & \textbf{Syntax errs (\%)} & \textbf{Location errs (\%)} & \textbf{Other (\%)} \\
\hline  
\rowcolor{gray!15}
\textbf{Ours} & 94{.}2 & 0 & 0 & 5{.}8  \\
DeepSeek R1 & 97{.}2 & 0 & 0{.}2 & 2{.}6 \\
GPT4 & 97{.}2 & 0 & 0{.}4 & 2{.}4 \\
Gemini & 98 & 0 & 2{.}0 & 0\\
Grok & 97 & 0 & 0 & 3 \\
\hline
\end{tabular}
\caption{Accuracy against state-of-the-art models}
\label{tab:results_summary}
\end{table}

Comparative analysis was conducted using the \emph{Semantic Variants} dataset subset, as this fragment represents the peak performance benchmark for our architecture. As summarized in Table~\ref{tab:results_summary}, our model achieved a 94.2\% exact match (EM) rate. Although this slightly trails the SOTA benchmark of 97.0\%, the results demonstrate that our solution maintains high accuracy and remains competitive within the current landscape of large-scale generative models.

\section{Application performance}\label{sec:app-performance}

\begin{figure}[h]
\centering
\begin{tabular}{cc}
\includegraphics[width=0.4\textwidth,trim=0 0 0 100, clip]{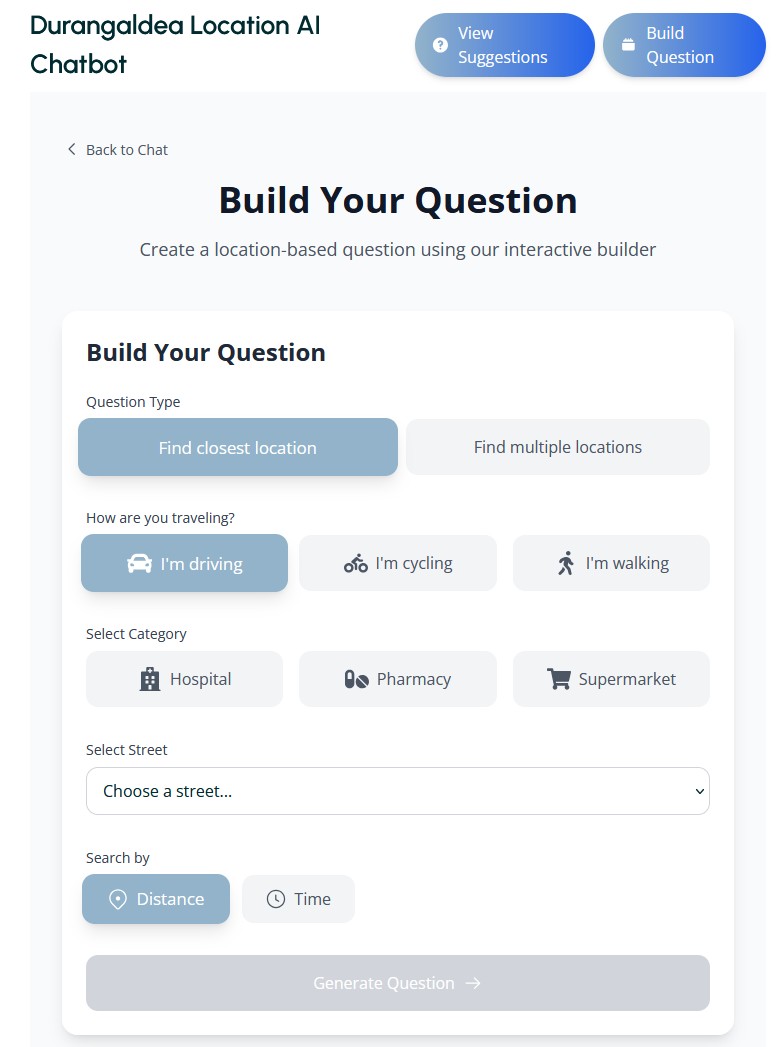} &
\includegraphics[width=0.4\textwidth]{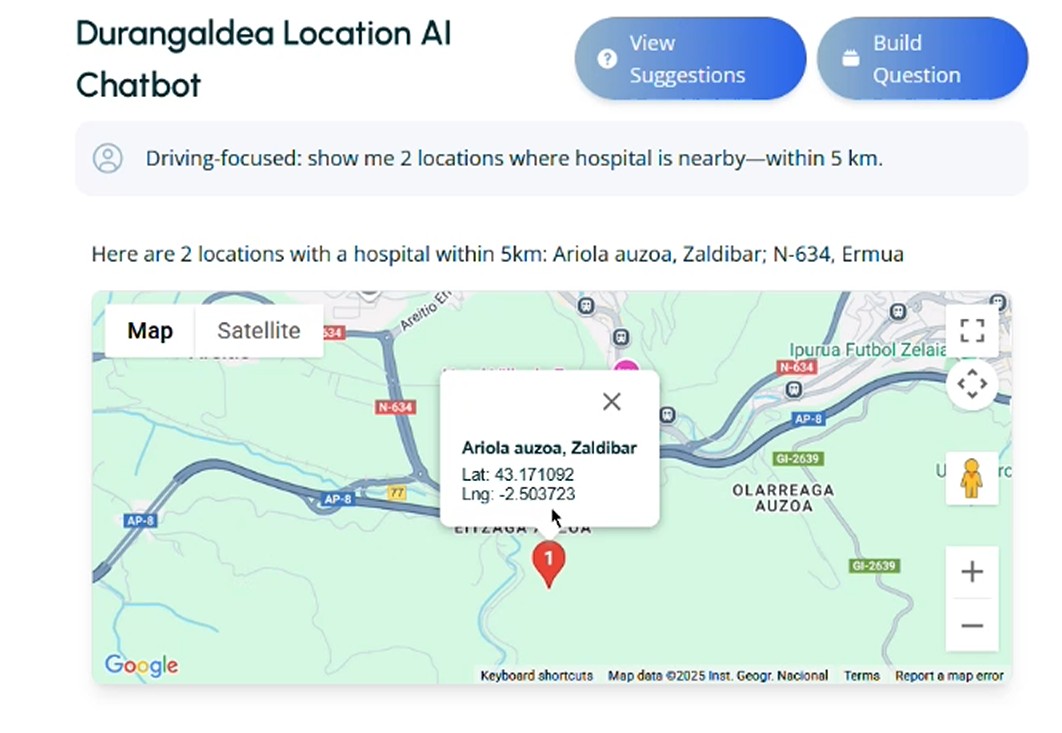}
\end{tabular}
\caption{Web application for model queries}\label{fig:app}
\end{figure}

The fine-tuned model was integrated into a web application that supports both free-form natural-language queries and on-click generation of predefined question types relevant to typical user needs. The system interfaces with Google Maps and relies on the Overpass API to convert street-level user inputs into geographic coordinates, enabling seamless interaction with the underlying dataset. Two representative screenshots are shown in Figure~\ref{fig:app}.

\begin{figure}[h]
\centering
\includegraphics[width=0.95\textwidth]{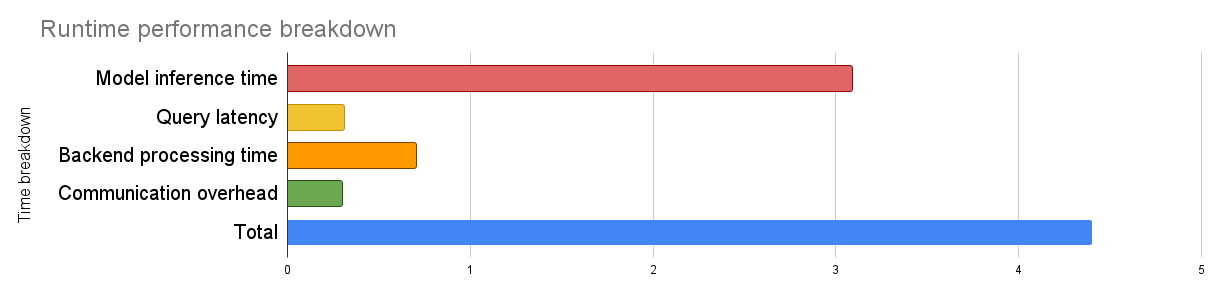} 
\caption{Application performance}\label{fig:app-performance}
\end{figure}

Model inference was performed on the same NVIDIA RTX 3090 GPU used during training. The averaged runtime breakdown in Figure~\ref{fig:app-performance} indicates that model inference is the primary source of latency, contributing roughly three seconds per request. In contrast, database lookup, backend logic, and communication overhead remain comparatively small.

Backend processing time also includes the application’s guardrail mechanisms, which filter and validate user inputs before forwarding them to the model. These guardrails ensure that the system rejects out-of-scope or unsafe queries, thereby reducing hallucinations, preventing unsupported queries, and enforcing domain constraints. Although this contributes modestly to backend latency, it plays a critical role in maintaining reliability and user safety.

The dominance of inference time suggests that end-to-end performance is chiefly limited by the model’s computational footprint. 

\section{Limitations \& Future work}\label{sec:limitations}

\textbf{Applicability to broader scenarios}. The application developed in this work is currently being integrated into the FUTURAL Metasearch platform, which provides unified access to heterogeneous agricultural data originating from multiple smart services. Our broader objective, however, is to establish a general methodology for connecting arbitrary queryable datasets to natural-language interfaces with minimal engineering and fine-tuning effort. A principal challenge in achieving this objective is the lack of pre-existing training data suitable for supervised fine-tuning. The synthetic data generation pipeline introduced in Section~\ref{sec:dataset} is a systematic solution to this problem: although not fully automated, it follows a principled procedure that can be adapted to datasets with similar structural characteristics.

Nevertheless, this methodology does not scale gracefully when a large number of datasets, each encoding distinct domain knowledge, are combined. In such settings, the space of admissible queries may grow exponentially, making manual template definition and projection-based question generation impractical. For these more complex, multi-domain scenarios, an approach grounded in ontology-based domain abstraction, coupled with LLM-driven topic and query-space generation, may offer significant advantages. Exploring such strategies is a promising direction which we consider for future research.


\textbf{Application scalability}. Our application is designed to process individual requests efficiently, but it does not yet scale to highly concurrent usage scenarios. This limitation is part of our design: the system was developed with  emphasis on simplicity and minimal hardware requirements rather than large-scale deployment. Nevertheless, several strategies can be employed to ensure scalability as demand grows. Some of these mechanisms are already being explored for the FUTURAL Metasearch platform. When latency constraints allow, multiple queries  can be grouped into a single forward pass through the model, a technique known as batching. Batching significantly improves GPU utilization and inference throughput, particularly for smaller models or bursty workloads. Additional optimizations—such as more aggressive quantization, low-rank adaptation methods, pruning, or using a smaller distilled model variant—can reduce inference time and memory consumption. These optimizations directly increase per-server capacity. Moreover, many user queries are expected to exhibit a relatively simple and repetitive structure, often involving distances, travel times, or nearest-location requests. As a result, numerous queries—or paraphrased variants of the same underlying question—can be served directly from a response cache rather than invoking the model for each request. 

\textbf{Graceful error recovery}
Our current architecture lacks automated recovery mechanisms for handling malformed or logically incorrect queries, relying instead on standard error reporting. To enhance system robustness, future iterations will explore iterative error correction strategies. This approach involves implementing a feedback loop where execution-time errors trigger a re-prompting sequence, enabling the model to refine and regenerate queries based on specific failure diagnostics.



\section{Related work}\label{sec:related-work}
The paper that introduces the means to connect a LLM with external APIs is \textbf{Toolformer}~\cite{Schick2023ToolformerLM}. Toolformer is a self-supervised framework that enables language models to learn how and when to call external tools—such as calculators, search engines, or APIs—using only minimal human annotation. The key idea is to let a pretrained model generate candidate tool calls within unlabeled text, execute those calls, and retain only the examples where the returned results measurably improve the model’s ability to predict the original text. These filtered examples form an augmented training set that teaches the model to invoke tools autonomously, integrate tool outputs into its reasoning, and produce more accurate final answers. This approach allows Toolformer to extend a model’s capabilities without requiring large-scale manual datasets or task-specific engineering.

Toolformer cannot operate directly on a database for which no training dataset exists because its learning process depends on having text in which the model can propose, execute, and verify tool calls. The method works only when the model can generate candidate queries in context, run them on the tool, and then measure whether the tool’s output improves its ability to predict the surrounding text. If no dataset contains references to the database’s contents—no examples of the entities, relationships, or queries associated with it.

\textbf{Gorilla}~\cite{Gorilla} is a large language model designed specifically for reliable API invocation at scale. It combines LLM generation with retrieval-augmented grounding, using API documentation and signatures to guide the model toward valid, executable calls. Rather than memorizing APIs, Gorilla retrieves relevant function specifications at inference time and conditions the model on them, enabling accurate argument selection and reducing hallucinated or invalid calls. This approach allows Gorilla to generalize to thousands of real-world APIs, making it one of the first LLMs explicitly optimized for large-scale, real-world tool use. 

In contrast, our method is intentionally lightweight and domain-specific, focusing on a single structured dataset with a fixed schema and a well-defined space of admissible queries. Rather than relying on large-scale API documentation or retrieval systems, the approach uses a synthetic dataset-generation pipeline to produce training examples tailored directly to the underlying database. This enables effective fine-tuning even when no pre-existing training corpus exists. Because the model is optimized for a narrow set of tasks, it avoids the complexity and overhead associated with multi-tool reasoning, leading to higher reliability within its domain.

Overall, Gorilla and our method occupy different points in the design space of tool-using language models. Gorilla excels when broad coverage across tools and domains is essential, while our approach is better suited for targeted, high-precision applications with minimal hardware and engineering overhead. The simplicity and reliability of the latter make it a practical solution for real-world systems where the domain is well defined and the cost of large-scale retrieval infrastructures cannot be justified.

\textbf{ReAct}~\cite{yao2022react} is a framework in which a language model interleaves natural-language reasoning steps with actions, such as search queries, tool calls, or environment interactions. Instead of producing a final answer directly, the model generates a sequence of thought–action–observation triplets: it reasons in text, issues a tool call, observes the tool’s output, and continues reasoning. This iterative loop enables the model to solve complex tasks that require external information, multi-step planning, or verification. Compared with our approach, ReAct is general-purpose and interactive, whereas our method is domain-specific and single-step. ReAct is designed for environments where the model must repeatedly query external tools and refine its reasoning, relying on dynamic feedback from the environment. By contrast, our system requires only one structured API call per query, generated directly from the natural-language input without multi-step planning. Moreover, ReAct assumes access to a rich tool ecosystem and depends on the model's ability to reason and act iteratively, which increases complexity and computational cost. Our approach focuses on simplicity, low hardware requirements, and high reliability within a fixed domain, achieved through a synthetic training pipeline tailored to the dataset’s structure. ReAct and Gorilla belong to a broader line of research exploring how language models interact with external tools. 

Earlier efforts such as 
\textbf{ToolAlpaca}~\cite{tang2023toolalpacageneralizedtoollearning} focus on training models to use tools through supervised demonstrations, but they depend heavily on manually curated instruction–tool pairs and therefore require substantial annotation effort. \textbf{Confucius}~\cite{gao2023confuciusiterativetoollearning} extends this idea by introducing iterative tool learning, where the model refines tool-use behaviors across multiple steps, but its multi-round interaction paradigm is computationally expensive and less suitable for low-resource environments. \textbf{GPT4Tools}~\cite{yang2023gpt4toolsteachinglargelanguage} demonstrates strong tool-use performance by leveraging a very large base model (GPT-4), yet its reliance on massive proprietary training data makes the approach difficult to reproduce in constrained academic or deployment settings. Finally, \textbf{CREATOR}~\cite{qian2024creatortoolcreationdisentangling} explores disentangling tool creation from tool usage, enabling language models to propose new tools entirely from descriptions; however, its scope is far broader than required for single-dataset applications and introduces significant model and system complexity.

In contrast to these general-purpose and often resource-intensive approaches, our method targets a specific structured dataset, requires minimal hardware, and avoids the need for large curated corpora or iterative interaction loops. By relying on a principled but lightweight synthetic dataset generation pipeline, our system achieves high reliability in a focused domain without the overhead associated with multi-tool or multi-step reasoning frameworks.

Finally, we mention Model Context Protocol (MCP)~\cite{anthropic_mcp_2024}, which has emerged as the de-facto standard for integrating external tools with large language models. MCP provides a uniform mechanism through which servers expose tools, each accompanied by a formally defined JSON schema that the client loads into the model’s working context. In principle, this offers a clean, typed interface for tool invocation. In practice, however, MCP places a substantial cognitive and computational burden on the model: a single MCP server can expose dozens of tools, collectively amounting to tens of thousands of tokens of schemas and descriptions that must be loaded into the context window before any task-specific reasoning begins. This leads to context saturation, competition between global tool definitions and task-relevant information, and a rapid degradation of tool-call accuracy—particularly in multi-step workflows, where small per-call error rates compound exponentially. These limitations highlight a core challenge: MCP exposes not only tools, but too much tool metadata, forcing the model to reason, plan, and select actions while navigating an unnecessarily large tool universe.

By contrast, our approach operates in a constrained, dataset-specific setting, avoiding the overhead of global schema loading and eliminating the need for the model to sift through irrelevant tools. Instead of exposing a large, heterogeneous action space, we train the model to produce a single, domain-specific API call, supported by a carefully constructed synthetic dataset. This results in significantly reduced hardware requirements, and higher reliability within the targeted domain.

\section{Conclusion}\label{sec:conclusion}
This paper presents a fully open-source, natural language question-answering (QA) framework optimized for accessibility data within the Durangaldea region of Spain. Our results demonstrate that the proposed system achieves performance metrics comparable to state-of-the-art (SOTA) large language models, while at the same time running on commodity hardware. It successfully bridges the gap between specialized local datasets and high-fidelity generative responses. Beyond the specific implementation, we propose a generalized methodology for the rapid deployment of QA systems across diverse tabular or relational datasets. By leveraging high levels of automation, this workflow significantly reduces the requirement for human expert intervention during the dataset preparation and model alignment phases. While the current architecture is optimized for a single-domain dataset, future research will focus on extending this system to multi-dataset environments, addressing the challenges of cross-domain query generation and data fusion to provide a more holistic analytical tool for arbitrary collections of datasets.

\begin{credits}
\subsubsection{\ackname} This paper has been funded by the by the European Union, via the FUTURAL project - \emph{Empowering the FUTure through innovative Smart Solutions for rURAL areas} (HORIZON EUROPE) Project ID 101083958. Views and opinions expressed are however those of the author(s) only and do not necessarily reflect those of the European Union or the European Research Executive Agency. Neither the European Union nor the granting authority can be held responsible for them.

\end{credits}
%
%
%
\bibliographystyle{splncs04}
\bibliography{bibliography}

\end{document}